\documentclass[11pt]{article}

\usepackage[final]{acl}

\usepackage{times}
\usepackage{latexsym}

\usepackage[T1]{fontenc}

\usepackage[utf8]{inputenc}

\usepackage{microtype}

\usepackage{inconsolata}

\usepackage{graphicx}
\usepackage{caption}
\usepackage{booktabs}
\usepackage{amsmath}
\usepackage[table]{xcolor}
\usepackage{xcolor}
\usepackage{makecell}
\usepackage{subcaption}

\title{Fast and Accurate Quotation Attribution in Literary Texts}

\author{Gaspard Michel\textsuperscript{\normalfont{1,2}},  \;\; Hugo Attali\textsuperscript{\normalfont{3}}, \;\; Elena V. Epure\textsuperscript{\normalfont{1,4}}
\\
\\
{%
\normalsize
\begin{tabular}{@{}l@{}}
  \textsuperscript{\normalfont{1}}{\normalfont{Deezer Research, Paris, France}} \\
  \textsuperscript{\normalfont{2}}{\normalfont{LORIA, Nancy, France}} \\
  \textsuperscript{3}{\normalfont{Université Sorbonne Paris Nord, CNRS, LIPN}} \\
  \textsuperscript{4}{\normalfont{IDIAP, Martigny, Switzerland}}
\end{tabular}
}%
\\
\\
  \normalfont{\texttt{gaspard.michel@loria.fr}}
}

\begin{document}
\maketitle
\begin{abstract}
Attributing quotations to their speakers in literary texts remains an open challenge. Standard methods, which independently predict a speaker mention for each quotation, are efficient but still limited in accuracy. In contrast, large language model (LLM) approaches achieve strong performance, but their computational cost limits their use in large-scale literary analysis. We propose an encoder-based efficient formulation that resolves multiple quotation attributions within a shared, large context window. 
Using our new formulation, \textit{joint scoring}, we report state-of-the-art (SOTA) performance on the Project Dialogism Novel Corpus (PDNC), comprising more than 35,000 manually annotated quotations from 22 English novels.
Our best model reaches 94.5\% overall attribution accuracy while processing novels $20\times$ faster than comparable standard methods and more than $1000\times$ faster than LLM-based approaches on an A100 GPU.
An analysis of models' representations suggests that joint scoring improves on challenging attribution examples by preserving long-range anaphora resolution signal, an information that we found already present in pretrained encoders.
To facilitate adoption, we release ModernBookNLP, a modified fork of BookNLP that replaces its quotation attribution model with our best system available at \url{https://github.com/gasmichel/ModernBookNLP_QA/}.
\end{abstract}

\section{Introduction}

The automatic attribution of quotations to their speakers in literary texts, also known as \textit{quotation attribution}, is crucial for a range of large-scale literary analyses, including \textit{character network extraction} \cite{Elson2010, Muzny2017a, Labatut2019, sims-bamman-2020-measuring}, \textit{emotion arc analysis} \cite{vishnubhotla-etal-2024-emotion}, and \textit{character voice analysis} \cite{michel-etal-2024-distinguishing}.
In practice, quotation attribution identifies a character mention, either named, pronominal, or nominal, as the speaker of a target quotation, using a broader text segment containing the quotation as context. 
This process is known as \textit{quote--mention linking}. 
When the selected mention is pronominal or nominal, an additional \textit{mention--character linking} step is required to associate the mention with a character identifier (\textit{e.g.} ``he'' $\xrightarrow[]{}$ MR\_BENNET), thus resolving the relevant coreference relation \cite{Elson2010a}.

Existing systems typically build supervised quote-mention linking systems, combined with mention-character linking that rely on off-the-shelf coreference resolution to link pronominal and nominal mentions to character identifiers \cite{Vishnubhotla2023, michel-etal-2024-improving}, or mention-character linking that restrict mentions to unambiguous named mentions only \cite{zhong2025simple, michel2026graphlitlearningtextenricheddynamic}.
However, quotation attribution remains a challenging task, and recent approaches, despite their efficiency, continue to face limitations in predictive accuracy.
Conversely, the current SOTA on English novels is achieved by the LLM-based approach of \citet{michel-etal-2025-evaluating}.
While this solution achieves strong accuracy, it remains highly inefficient, thus limiting its applicability to very large corpora of literary texts.

In this work, we aim to provide the research community with a \textit{quotation attribution model} that is \textit{both fast and accurate}.
Inspired by the strong performance of LLM-based approaches and the efficiency of span-based, multi-instance task formalizations such as in coreference resolution \cite{lee-etal-2017-end}, we propose \textit{joint scoring}, an adaptation of this formalization to the quotation attribution task.
Joint scoring fine-tunes textual representation models to jointly resolve multiple attribution decisions within a single, shared contextual window.
This departs from standard approaches, which attribute each contextualized quotation independently, even when multiple quotations occur within the same or overlapping contexts.
We refer to these approaches as \textit{direct scoring}.
Through a systematic evaluation of key design choices, we compare the two scoring approaches across the choice of textual representation model, the amount of contextual information available, and the types of candidate mentions (pronominal, nominal, named) considered when performing quote–mention linking.
The evaluation is conducted on PDNC \cite{vishnubhotla-etal-2022-project} with more than 35,000 manually annotated quotations from 22 classic English novels.

Using ModernBERT \cite{warner-etal-2025-smarter} as the textual encoder, we report SOTA performance on PDNC for our joint scoring model, reaching a maximum overall attribution accuracy of $94.5\%$ and $92.4\%$ on challenging attribution examples (such as complex conversational settings) at the largest context window considered.
Besides, joint scoring processes novels 20$\times$ faster than direct scoring, at the cost of increased but manageable GPU VRAM usage, and more than 1000$\times$ faster than the LLM-based approach of \citet{michel-etal-2025-evaluating} on comparable inference infrastructures.
Holding the context size constant, we found that models using pronominal, nominal, and named mention candidates (\textit{coreference clusters}) systematically outperform systems that consider only named mentions (aliases), contradicting previous findings that identified coreference resolution as the main bottleneck \cite{ Vishnubhotla2023, michel-etal-2024-improving}.

Then, we carry out a thorough analysis of representations built by fine-tuned and pre-trained models.
We found that joint scoring's advantage over direct scoring stems from its ability to better preserve ModernBERT's initial ability to link two mentions referring to the same character (anaphora resolution) across long range.
Besides, we show that anaphora resolution signal is well correlated with downstream accuracy on complex attribution settings such as discussions and quotes referred to with pronouns (e.g. ``she said'').


To facilitate its adoption, we release a modified fork of BookNLP\footnote{\url{https://github.com/gasmichel/ModernBookNLP_QA/tree/main/ModernBookNLP}}, which replaces the original quotation attribution model with our best-performing system.
The modified pipeline substantially improves attribution performance on LitBank \cite{bamman-etal-2020-annotated, sims-bamman-2020-measuring}, a corpus of 100 book passages annotated with various literary information, while achieving a similar inference speed on a consumer-grade GPU despite having 4$\times$ as many model parameters.
We view these results as a practical step toward accurate automatic literary analysis without sacrificing efficiency. 
In short, we make the following contributions:
\begin{itemize}
    \item We propose a new formalization of the quotation attribution task inspired by the span-based joint scoring paradigm, showing that it significantly improves downstream performance and efficiency over previous SOTA (+4.7 overall accuracy, at least $1000\times$ faster) and over direct scoring (+2-3 overall accuracy, $5\times$ to $20\times$ faster).
    \item We identify several key factors that improve downstream quotation attribution accuracy, apart from joint scoring, including encoder capability, context-window size, and coreference-based mention candidates.
    \item We conduct a detailed analysis of fine-tuned and pretrained model representations to interpret joint scoring’s advantage over direct scoring. We show that it stems from its better preservation of initial ModernBERT’s long-range anaphora resolution signal.
    \item We open-source a modified fork of BookNLP which includes our best-scoring systems, and show substantial improvements in quotation attribution accuracy through an external validation that matches a real-world scenario.
\end{itemize}

\section{Related Work}
\label{sec:related_work}

\subsection{Quotation Attribution}

Quotation attribution has traditionally been approached with supervised system trained on manually defined features \cite{Elson2010a, He2013}, such as speaker alternation patterns and character-level features \cite{He2010}, deterministic sieves of hand-crafted rules  \citet{Muzny2017}, sequence labeling \cite{okeefe-etal-2012-sequence}, Dialogue State Tracking \cite{CuestaLazaro2022} on English novels, or scoring systems on Chinese novels \cite{Chen2021}.

Recently, a large body of work has approached quotation attribution by fine-tuning pretrained textual encoders such as BERT \cite{devlin-etal-2019-bert}.
BookNLP\footnote{\url{https://github.com/booknlp/booknlp}}, a widely used natural language processing pipeline designed for literary texts, features a BERT model fine-tuned on quotation-attribution 100 book passages from LitBank \cite{sims-bamman-2020-measuring}.
In this pipeline, quote-mention linking is predicted by the fine-tuned BERT model, while mention-speaker linking is handled by BookNLP's coreference resolution model.

This standard approach has since been extensively reused and extended.
\citet{Vishnubhotla2023} extended BookNLP with restrictive rules that avoid spurious coreference links, while \citet{michel-etal-2024-improving} showed that replacing BERT with SpanBERT \cite{joshi-etal-2020-spanbert} and using broader contextual information substantially yielded large improvements.
Other works have reported gains by removing coreference-based mention-speaker linking altogether,  considering only named mentions during quote-mention linking, along with the use of more capable representation models \cite{zhong2025simple, michel2026graphlitlearningtextenricheddynamic}, sometimes by fine-tuning decoder-only models to directly generate a speaker name \cite{Su2023}.
Despite their inference efficiency, these approaches remain limited in accuracy, especially on challenging cases such as long dialogues involving multiple speakers.

\begin{figure}[t!]
    \centering
    \includegraphics[width=\linewidth]{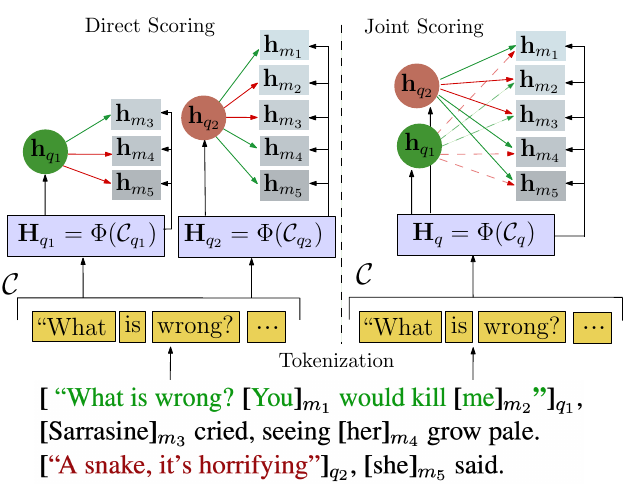}
    \caption{\textit{Left:} standard approaches score independent quotes against identified mentions in their context $\mathcal{C}$. \textit{Right:} we score jointly all quotations in $\mathcal{C}$ against identified mentions, with shared contextual embeddings $\mathbf{H}$. We let $\Phi$ be any textual encoder and use green arrows to indicate mentions referring to the gold speaker.}
    \label{fig:desc_fig}
\end{figure}

A common feature of these approaches is that they formalize quotation attribution as a \textit{direct scoring} task, that we summarize in Figure~\ref{fig:desc_fig} (left).
Namely, a single quotation $q$ is augmented with its surrounding context, yielding a contextualized segment $\mathcal{C}_q$, in which candidate character mentions have been initially identified (\textit{e.g.} $m_3=$``Sarrasine'' in Figure \ref{fig:desc_fig}). 
Then, given single quotation instances, a textual representation model $\Phi$ is fine-tuned to find all mentions referring to the quote-speaker (Sarrasine $m_3$, for the first quotation $q_1$).

In contrast, the LLM approach of \citet{michel-etal-2025-evaluating}---the current SOTA on PDNC---directly performs joint quotation attribution within very large context windows (\textit{i.e.} predicting all speakers of all quotations in a single shared context).
Its strong performance is attributed by the authors to the LLM's ability to better understand conversational patterns and perform local coreference resolution \citet{michel-etal-2025-evaluating}.
However, the reported processing times range from 10 minutes to an hour on a A100 GPU card, a substantial limitation for Digital Humanities practitioners working with large corpora of literary texts.

In the current work, we propose the first quotation attribution model that relies on span-based \textit{joint scoring} (Figure~\ref{fig:desc_fig}, right), a unification of the accurate but inefficient LLM-based joint attribution paradigm and the fast but less accurate direct scoring formalization.

\subsection{Joint Scoring in NLP}

Joint scoring is not new, and has been applied to several NLP tasks \citet{lee-etal-2017-end, yao-etal-2019-docred, he-etal-2018-jointly}.
Its core idea is to encode a large text segment with a representation model and resolve multiple task decisions from the same shared contextual embeddings.
This principle underlies the success of end-to-end neural coreference resolution.
\citet{lee-etal-2017-end} introduced end-to-end span-based scoring over shared contextual embeddings, which was later extended with higher-order iterative refinement \cite{lee-etal-2018-higher} and simplified by removing explicit span representations altogether \cite{kirstain-etal-2021-coreference}.
Similar shared-context formulations have proven effective in document-level relation extraction \cite{yao-etal-2019-docred}, semantic role labelling \cite{he-etal-2018-jointly}, and general information extraction tasks \cite{luan-etal-2019-general, wadden-etal-2019-entity}.
We adapt this principle to quotation attribution by building quotation and mention representations from the same encoded context and optimizing several attribution decisions jointly.
To our knowledge, this is the first application of joint scoring to quotation attribution.


\section{Problem Definition}
\label{sec:problem}

\paragraph{Direct Scoring.} Let $D = (t_1,\dots t_n)$ be a tokenized document and $q = (t_i, \dots, t_{j})$ be a quotation in this document, starting at token $i$ and ending at token $j$.
A contextual segment is computed for each quote, $ \mathcal{C}_q = [\, c_{q}^{left} \; | \, q \, | \; c_{q}^{right} \,]$, where $c_{q}^{left}$ and $c_{q}^{right}$ are left and right contextual information of length $\frac{T}{2}$ tokens and $[\, | \,]$ denotes concatenation.
Within each, $\mathcal{C}_q$, mention identification—using coreference resolution, NER, or alias matching—produces a candidate set of mentions $\mathcal{M}_q = [m_1,\dots,m_{n_q}]$ where $m_i=(t_{\text{start}(m_i)}, t_{\text{end}(m_i)}$) and start(.) and end(.) are indexing functions.
Each segment is fed into an encoder $\Phi$, producing a contextualized representation $\mathbf{H}_q = \Phi(\mathcal{C}_q)$, for each token in $\mathcal{C}_q$.
Then, quote representations are extracted by concatenating their start and end tokens: $\mathbf{h}_q =[\mathbf{H}[{t_i}]\;|\;\mathbf{H}[{t_j}]\,]$.
Similar representations are built for each candidate in $\mathcal{M}_q$, and an unary score is computed for each candidate by scoring quotations against candidates:

$$
s(q,m_i) = \psi([\mathbf{h}_q \; | \; \mathbf{h}_{m_i}])
$$
where $\psi$ is a multilayer perceptron (MLP).

Now, let $G(q) = \{m \in \mathcal{M}_q, ref(m)=spk(q)\}$ be the set of gold mentions referring to the speaker of a quote $q$, where $ref(m)$ and $spk(q)$ are functions assigning the referent character of a mention $m$ and the speaker of a quote $q$ respectively.
We maximise the likelihood of $G(q)$ with a negative log-likelihood loss:

{\small
\begin{equation*}
\mathcal{L}(q) = \log\sum\limits_{m \in \mathcal{M}_q} \exp(s(q,m)) - \log\sum\limits_{m \in G(q)} \exp(s(q,m))
\end{equation*}
}%

\noindent At inference time, the highest scoring mention is chosen as the speaker of the quotation.
\paragraph{Joint Scoring.} We divide each document in contextual segments $\mathcal{C}$ of length $T$ tokens, with a stride of $S$ tokens.
For each segment $\mathcal{C}$, we extract mention candidates $\mathcal{M}_{\mathcal{C}}$ and quotation candidates $\mathcal{Q}_{\mathcal{C}}$ that are located within $\mathcal{C}$.
Then, the joint scoring formalization follows the same extraction process to derive quotation and mention representations as above.
However, the loss is now defined as:

$$
\mathcal{L}_{joint}(\mathcal{C}) = \frac{1}{|\mathcal{Q}_{\mathcal{C}}|}\sum\limits_{q \in \mathcal{Q_{\mathcal{C}}}} \mathcal{L}(q)
$$
Thus, in each data instance, different quotation and mention representations are built from a shared contextual representation $\mathbf{H}_{\mathcal{C}}$, and are \textit{jointly} optimized via direct scoring.
Joint scoring improves efficiency because several attributions share one encoder forward pass.
Its multi-instance objective may also encourage local coreference structure, since quotations uttered by different speakers are optimized against mentions referring to these speakers within the same contextual representation, thus maximizing the likelihood of coreferent clusters $\{G(q), \; q \in \mathcal{Q}_\mathcal{C}\}$.
We evaluate this hypothesis in Section~\ref{sec:rsa}.


\section{PDNC Experiments}

\begin{table*}[ht]
\centering
\small
\renewcommand{\arraystretch}{0.95}
\setlength{\tabcolsep}{3.25pt}
\begin{tabular}{l c c c c |c c c |c c}
\toprule
Model & (T,S) & $M_{type}$ & Overall & Non-Explicit & Explicit & Anaphoric & Implicit & \makecell[c]{{Book Time (s)}} & \makecell{Peak VRAM} \\
\midrule
\rowcolor{gray!10}  Llama-3-8b & (4096, 1024) & Alias & 89.8\tiny{$\pm$2.7} & 87.4\tiny{$\pm$3.2} & 94.9\tiny{$\pm$1.1} & 93.1\tiny{$\pm$1.2} & 81.8\tiny{$\pm$5.7} & [400,2800] & 14GB \\
\rowcolor{gray!10} CEQA & (1000, 0) & Alias & 82.6 & 74.9 & 98.5 & 75.3 & 73.1 & -- & -- \\
\rowcolor{gray!10}  BookNLP+ & (200, 0) & Coref & 78.5\tiny{$\pm$4.0} & 68.9\tiny{$\pm$4.4} & 98.6\tiny{$\pm$1.2} & 70.2\tiny{$\pm$7.0} & 66.4\tiny{$\pm$5.7} & -- & -- \\
\rowcolor{gray!10}  GraphLit & (500, 0) & NER & 86.5\tiny{$\pm$1.7} & 83.3{\tiny $\pm$1.4} & 93.0{\tiny $\pm$3.0} & 85.9{\tiny $\pm$2.7} & 80.4{\tiny $\pm$4.0} & -- & -- \\
\midrule
\texttt{Direct} & (1000, 256) & Alias & 83.3\tiny{$\pm$6.6} & 77.0\tiny{$\pm$7.4} & 96.2\tiny{$\pm$4.4} & 78.8\tiny{$\pm$8.3} & 74.0\tiny{$\pm$7.7} & 5.16\tiny{$\pm$2.5} & 2.4GB \\
\texttt{Joint} & (1000, 256) & Alias & 85.0\tiny{$\pm$6.7} & 79.6\tiny{$\pm$7.6} & 96.2\tiny{$\pm$4.5} & 81.5\tiny{$\pm$9.6} & 76.7\tiny{$\pm$7.8} & 0.39\tiny{$\pm$0.2} & 2.5GB \\
\midrule
\texttt{Direct} & (500, 256) & Coref & 91.2\tiny{$\pm$3.0} & 87.8\tiny{$\pm$3.3} & 98.6\tiny{$\pm$1.2} & 88.9\tiny{$\pm$4.2} & 85.1\tiny{$\pm$4.6} & 3.28\tiny{$\pm$1.6} & 2.1GB \\
\texttt{Direct} & (1000, 256) & Coref & 92.3\tiny{$\pm$3.0} & 89.3\tiny{$\pm$3.5} & 98.8\tiny{$\pm$0.9} & 91.6\tiny{$\pm$2.9} & 86.6\tiny{$\pm$4.3} & 5.38\tiny{$\pm$2.7} & 2.4GB \\
\texttt{Direct} & (2000, 512) & Coref & 92.6\tiny{$\pm$2.5} & 89.7\tiny{$\pm$3.0} & 98.9\tiny{$\pm$0.9} & 91.6\tiny{$\pm$3.0} & 86.9\tiny{$\pm$3.6} & 10.81\tiny{$\pm$5.4} & 2.9GB \\

\midrule
\rowcolor{green!10} \texttt{Joint} & (500, 256) & Coref & 92.7\tiny{$\pm$2.6} & 89.9\tiny{$\pm$2.9} & 98.8\tiny{$\pm$1.2} & 91.6\tiny{$\pm$3.6} & 86.9\tiny{$\pm$5.0} & 0.68\tiny{$\pm$0.3} & 2.5GB \\
\rowcolor{green!10} \texttt{Joint} & (1000, 256) & Coref & 93.5\tiny{$\pm$2.4} & 91.0\tiny{$\pm$2.8} & 98.9\tiny{$\pm$0.8} & 93.2\tiny{$\pm$2.3} & 88.2\tiny{$\pm$4.0} & \textbf{0.41}\tiny{$\pm$0.2} & 4.1GB \\
\rowcolor{green!10} \texttt{Joint} & (2000, 512) & Coref & \textbf{94.5}\tiny{$\pm$2.4} & \textbf{92.4}\tiny{$\pm$2.8} & \textbf{99.3}\tiny{$\pm$0.7} & \textbf{95.5}\tiny{$\pm$2.0} & \textbf{89.3}\tiny{$\pm$3.5} & 0.43\tiny{$\pm$0.2} & 9.2GB \\
\midrule
\texttt{D-}\scriptsize{Longformer}& (2000, 512) & Coref & 86.0\tiny{$\pm$4.2} & 80.6\tiny{$\pm$4.6} & {97.6}\tiny{$\pm$1.7} & 83.6\tiny{$\pm$4.3} & 77.4\tiny{$\pm$3.2} & 20.5\tiny{$\pm$9.7} & 5.4GB \\
\texttt{J}-\scriptsize{Longformer}& (2000, 512) & Coref &  {87.7}\tiny{$\pm$3.6} & {83.3}\tiny{$\pm$4.0} & 97.3\tiny{$\pm$1.1} & {87.4}\tiny{$\pm$3.6} & {79.5}\tiny{$\pm$2.9} & 0.82\tiny{$\pm$0.4} & 6.4GB \\
\bottomrule
\end{tabular}
\caption{Quotation attribution mean accuracy across folds (standard deviation in parentheses) on PDNC across context sizes ($T$), mention types ($M_{type}$) ; and inference statistics---average forward pass time in seconds of entire PDNC novels and peak GPU VRAM on a A100 GPU. Best results are bolded; grey rows indicate reported baseline results. Our methods use \texttt{ModernBERT-large}, except for bottom rows that use \texttt{Longformer-base}. Joint scoring (green) consistently outperforms other methods while being faster.}
\label{tab:results}
\end{table*}

\begin{figure*}[t]
    \centering
    \includegraphics[width=\linewidth]{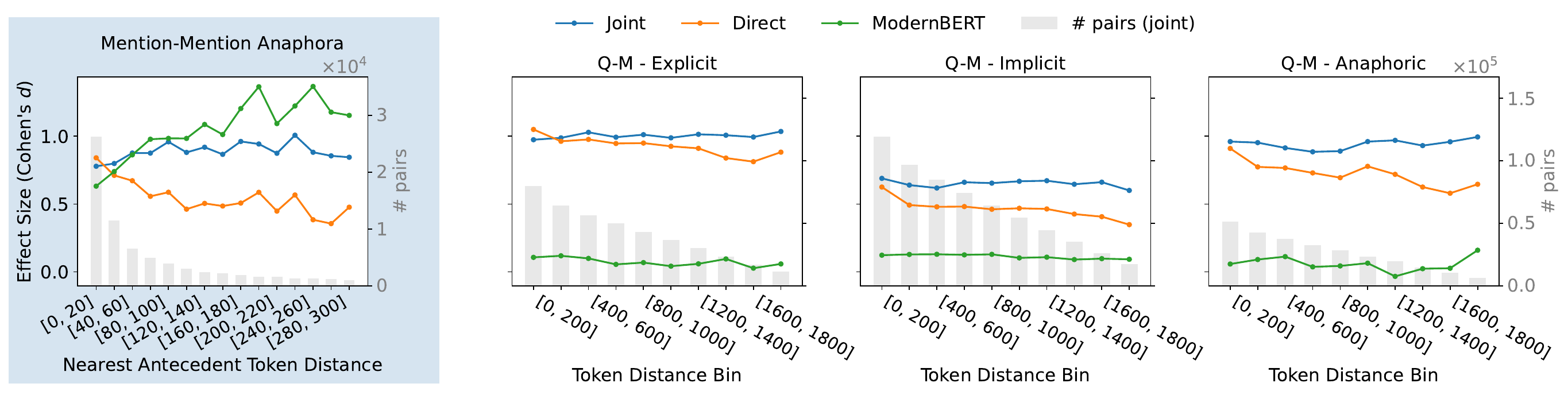}
    \label{fig:quote-quote}


    \includegraphics[width=\linewidth]{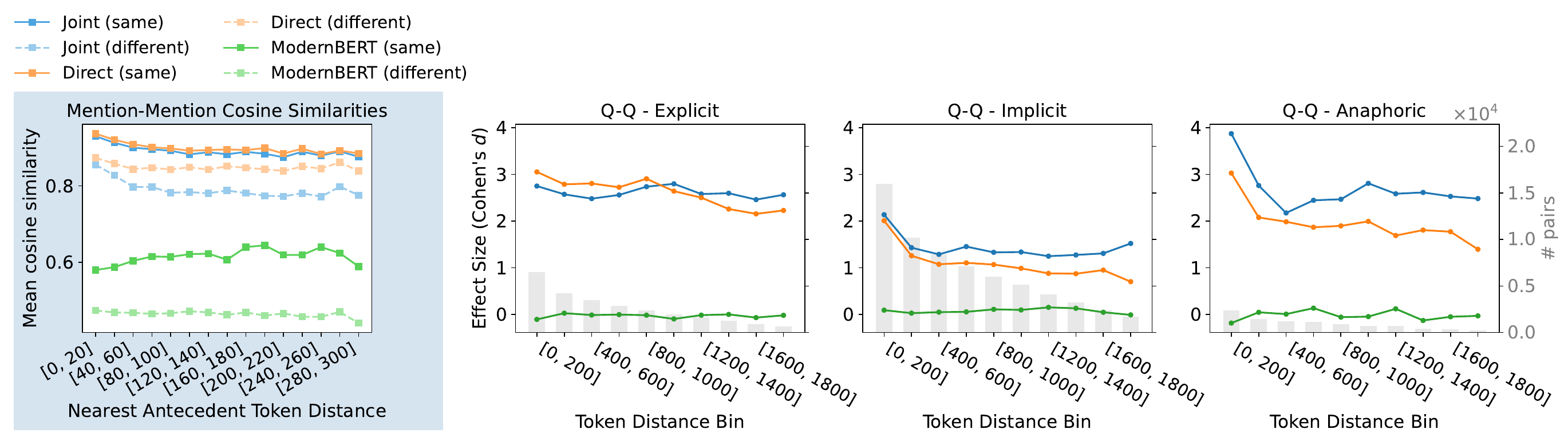}
    \label{fig:quote-quote}

    \caption{\textit{Left column}: Mention-Mention same-/different-character effect sizes (Cohen's $d$, top) and mean cosine similarities (bottom), as a function of distance to the nearest antecedent mention. \textit{Right panels}: Cohen's $d$ effect sizes for same- vs. different-speaker cosine similarity between Quotation-Mention (Q-M, top row) and Quotation-Quotation (Q-Q, bottom row) representations, by quotation type and token distance. Grey bars indicate pair counts per bin (right axis). Joint maintains higher separation than Direct at larger distances, especially for Anaphoric/Implicit quotations and Mention-Mention pairs, while ModernBERT shows negligible Q-Q/Q-M separation but a distinct, increasing native discriminative signal for Mention-Mention (anaphora) resolution.}
    \label{fig:representation-analysis}
\end{figure*}

Throughout this section, we use PDNC's 5-fold cross-validation, which divides the corpus at the book level.
Each quotation is assigned a \textit{type} based on how its speaker is introduced: \textit{explicit} quotations are introduced by a named mention (``Sarrasine cried''), \textit{anaphoric} quotations by a pronominal or nominal mention (``she said''), and speakers of \textit{implicit} quotations are not introduced by the narrator, often occurring in conversational settings.
For coreference-based models, following standard practice, we apply BookNLP to each novel and retain only clusters resolvable to a single character in the annotated gold PDNC lists and report scores for characters mentioned at least 10 times \cite{Vishnubhotla2023, michel-etal-2024-improving}.
For models that use aliases, we restrict coreference chains to named mentions only, which we described in details in Appendix~\ref{app:coref}.

\paragraph{Baselines} On PDNC, we compare against three direct scoring systems, CEQA \cite{zhong2025simple}, BookNLP+ \cite{michel-etal-2024-improving} and GraphLit \cite{michel2026graphlitlearningtextenricheddynamic}, and the joint scoring LLM-based approach \cite{michel-etal-2025-evaluating} with Llama-3-8b (8-bits quantized with the \textit{BitsAndBytes} library), with greedy decoding. 
Each model is evaluated under a similar evaluation protocol, accessing the gold-labelled character list both at training and evaluation time.
Besides, differently from \citet{michel-etal-2025-evaluating}, we report Llama-3-8b results under the exact 5-fold cross-validation used to evaluate other models, and report processing times from our inference infrastructures.
Thus, reported results across models are directly comparable.

\paragraph{Evaluation} 
We report average cross-validation accuracy over the 5 standard PDNC splits, and gather per-book inference statistics to evaluate efficiency averaged across 5 runs, where we set the batch size of each model to 16.
These include the average processing time of PDNC novels in terms of forward pass time in seconds, and peak GPU VRAM, but it does not entail upstream BookNLP inference (tokenization, named-entity-recognition and coreference resolution), since we regard this as a preprocessing step, for which each tested method has access to at inference time.
All models were trained and evaluated on a single A100 GPU, and use Flash Attention \cite{dao2022flashattentionfastmemoryefficientexact} with \textit{bfloat16}.
A notable downside of upstream coreference resolution is that predicted coreferent clusters can be noisy, and sometimes lead to quotations to have no mention candidates referring to the ground-truth speaker, even in very large contexts (2\% of PDNC quotations when $T=2000$).
We remove such quotations during training but we consider them at evaluation time, ensuring a realistic setup.

\subsection{Quantitative Results}
\label{sec:quant}

In what follows, when \texttt{ModernBERT-large} is used as an encoder, we report four model configurations: direct and joint scoring, each with aliases or coreference information.
Models that use aliases only keep named-mention candidates during training and inference, while models that use coreference information also include pronominal and nominal mentions.
Further, after establishing the superiority of coreference-based systems, we evaluate the impact of context length using models trained with different values of context length $T$ and stride $S$.
Finally, we report the results for our best configuration when swapping the encoder to \texttt{Longformer-base}.

Training and inference details, along with additional results on minor PDNC characters, are provided in Appendix~\ref{app:app_1}.
We display evaluation results in Table~\ref{tab:results}, and discuss our main findings below.

\paragraph{Joint scoring is accurate and efficient.} Across comparable context lengths and mention types, joint scoring systematically outperforms direct scoring while being $4\times$ faster with $T=500$ and $20\times$ faster with $T=2000$ (McNemar $p<0.001$ except for Explicit with $T=1000$). Notably, joint scoring with $T=2000$ achieves SOTA results on PDNC, improving overall accuracy by $5$ points over the previous best-performing Llama-3-8b model while being $1000\times$ faster. This highlights the scalability of joint scoring: dividing long novels into larger contexts reduces the number of encoder forward passes, enabling efficient processing but with a higher, but still manageable GPU VRAM usage\footnote{Smaller batches trade efficiency for lower VRAM usage.}.

\paragraph{Coreference-based systems are more accurate.} Compared with direct and joint scoring systems using aliases ($M_{type} = $ Alias), similar coreference-based systems ($M_{type} = $ Coref) systematically improve downstream attribution accuracy when tested with $T=1000$ and $S=256$. Gains are notably larger for non-explicit quotations, reaching improvements of 12 accuracy points for both direct and joint. Contrary to prior work which used small context sizes,
\cite{michel-etal-2024-improving}, coreference is found to play a key role in accurate quotation attribution when accessing larger contexts.

\paragraph{Larger context lengths yield better performance.} For both direct and joint scoring, increasing context length yields small but significant improvements in non-explicit attribution accuracy (McNemar $p<0.001$ within the same model and paradigm). Models using the largest context ($T=2000$) systematically improve non-explicit accuracy by around $2$ points over those using $T=500$ tokens.

\paragraph{Results hold for less capable encoders} 
When using Longformer-base as the base encoder in our model, which has $3\times$ less parameters than ModernBERT-large, we observe similar findings when comparing joint and direct scoring.
Indeed, joint scoring still shows improved performance over direct scoring on non-explicit quotations (McNemar $p<0.001$ except for Explicit quotations), while being more efficient. Interestingly, we found that for similar context sizes and scoring task, Longformer-base inference times are around twice as slow as ModernBERT-large, which can be explained by its lack of support for Flash-Attention.

\subsection{Representation Similarity Analysis}
\label{sec:rsa}

Having established that joint scoring outperforms direct scoring in both speed and accuracy, this section examines the underlying factors that may account for this advantage.
We formulate different hypothesis, and conduct a representation similarity analysis on the test-set of the fourth fold of PDNC corpus, across three models: joint scoring, direct scoring both with $T=2000$ and $S=512$, and standard pre-trained ModernBookNLP.
We use models trained on the train set of the associated fold, ensuring that tested books were not seen at training time.
We also ensure that each model is inputted exactly the same context $\mathcal{C}$ when extracting downstream representations.
Below, we state each hypothesis, describe the experimental setup, and present the results (summarized in Figure~\ref{fig:representation-analysis}).

\paragraph{H1.} \textit{Under joint scoring, representations of quotations uttered by the same speaker within a shared context $\mathcal{C}$ become more similar to each other than under direct training.}

To test this hypothesis, we analyse quotation-quotation representation similarities. We extract quotation representations $\mathbf{h}_q$
 for all quotations within a shared context window $\mathcal{C}$. Then, given an anchor quotation $q$, we form \textit{positive pairs} with other quotations uttered by the same speaker, and \textit{negative pairs} with those uttered by different speakers, and bin each pair by the absolute token distance between the two quotations, using intervals of 200 tokens. Within each distance bin, we compute the cosine similarity of positive and negative pairs, and summarize their separation using Cohen's $d$, which quantifies how discriminable same-speaker quotations are from different-speaker quotations at that distance. We repeat this procedure for the three quotation types: Explicit, Implicit, and Anaphoric.

The Q-Q panels of Figure~\ref{fig:representation-analysis} (bottom right) show that under joint scoring, same-speaker quotation pairs are found more separable from different-speaker pairs than under direct scoring, particularly for Implicit and, to a lesser extent, Anaphoric quotations.
This advantage notably persists and widens as token distance increases.
For Explicit quotations, where the speaker is locally introduced by the narrator, direct and joint scoring achieve comparably high separation throughout, indicating that joint scoring's benefit is concentrated on complex attribution decisions, consistent with the implicit and anaphoric accuracy gains from Table~\ref{tab:results}.

\paragraph{H2.} \textit{Under joint scoring, representations of a quotation and mentions referring to their speaker within a shared context $\mathcal{C}$ become more similar to each other than under direct training.}

Here, we analyse quotation-mention representation similarities, derived using the same protocol as H1.
The only difference lies in how we build positive and negative pairs: for each quotation $q$, we form \textit{positive pairs} with mentions referring to $q$'s gold speaker, and \textit{negative pairs} with all mentions referring to a different character.

The Q-M panels (Figure~\ref{fig:representation-analysis}, top-right) show that joint scoring's separation between a quotation and its ground-truth speaker's mentions degrades far less with distance than under direct scoring, especially for Anaphoric quotations, matching exactly the category where Table~\ref{tab:results} shows the largest accuracy gap across comparable configurations ($+3.9$ points for $T=2000, S=512$).
Same-/different-character similarities (Figure~\ref{fig:representation-analysis}, bottom-left) shows that direct scoring's \textit{different}-character similarity increases with distance, gradually approaching same-character similarity, while joint scoring keeps different-character mentions comparatively well separated.
Joint scoring's advantage thus seems to be through preserving negative-pair discriminability at longer range rather than strengthening same-character consistency.

Intuitively, a model showing larger Q-M separability assigns larger similarities between quotation-mention pairs that refer to the same character/speaker, a signal likely to be correlated with downstream quotation attribution.
We analyse quantitatively if a correlation exists between per-quote-type accuracy and Q-M separability in Appendix~\ref{app:corr_qa}.
We observe a strong, highly significant positive relationship (Spearman $\rho=0.85$), between Q-M separability and overall quotation attribution accuracy, where highest correlations are observed for Anaphoric quotations ($\rho=0.8$) and Implicit quotations  ($\rho=0.67$).
Thus, we attribute the advantage of joint scoring over direct scoring on downstream quotation attribution through a better Q-M separation, particularly on Anaphoric and Implicit quotations which show both a stronger separability and accuracy gains.

\begin{table}[t!]
\centering

\setlength{\tabcolsep}{3.5pt}
\begin{tabular}{l c c c}
\toprule

& QA& Coref & Time (s)\\
\midrule
\texttt{BookNLP}       & 68.1 & 71.3 & \textbf{5.2} \tiny{$\pm$ 0.4} \\
\texttt{ModernBookNLP} \scriptsize{(Direct)} & 72.6 & 71.4 & 10.8 \tiny{$\pm$ 4.8} \\
\texttt{ModernBookNLP} \scriptsize{(Joint)} & \textbf{77.5} & {71.5} & 5.5 \tiny{$\pm$ 0.4} \\
\bottomrule
\end{tabular}
\caption{CoNLL F1 scores on LitBank (QA: quotation attribution; Coref: coreference resolution) and average processing time per book (in seconds) on a Quadro RTX 6000 without FlashAttention or bfloat16 support.}
\label{tab:booknlp_conll_time}
\end{table}

\paragraph{H3.} \textit{Under joint scoring, representations of all mentions referring to the same character show better clustering than under direct scoring.}

Here, we analyse mention-mention representation similarities through anaphora, a signal of genuine coreferent clustering.
For each mention $m$, we identify the \textit{nearest preceding mention}--often called anaphora--referring to the same character within $\mathcal{C}$ and form a \textit{positive pair} $(m, m')$ binned by their absolute token distance, using intervals of 20 tokens up to a maximum distance of 300 tokens.
To keep a comparable number of positive and \textit{negative pairs}, we randomly sample $k=2$ mentions per distinct character within the same distance bin (the same sampling is applied per model), which avoids over-representing frequently mentioned characters.
We report per-bin raw mean cosine similarity of positive and negative pairs, and their separation as measured by Cohen's $d$ in Figure \ref{fig:representation-analysis}, left column.

Interestingly, we found that anaphora resolution signal--the ability to link two coreferring mentions together--is naturally encoded in ModernBERT-large while showing negligible Q-Q/Q-M separation, indicating it cannot naturally distinguish same- vs different-speaker quotations and speaker mentions.
This discriminative signal on Mention-Mention pairs increases as distance grows, surpassing direct scoring at longer ranges  (Figure~\ref{fig:representation-analysis}, top-left).
This suggests pretraining alone already encodes substantial coreference-relevant structure \cite{clark-etal-2019-bert}, and that direct scoring's per-instance fine-tuning objective degrades part of this long-range structure to instead focus on small-range interactions, where it shows the highest separability across models.
These results are consistent with the representation collapse observed more broadly in fine-tuned encoders \cite{ethayarajh-2019-contextual, gao2018representation}, where already solid capabilities of pretrained-models are eroded through task-specific fine-tuning.
In contrast, joint scoring's shared-context, multi-instance objective better preserve this pretrained structure, offering one plausible explanation for why its accuracy advantage over direct scoring grows with context size.

\section{External Validity on LitBank}

Standard PDNC evaluation uses gold-labelled list of character aliases, a scenario often unrealistic when processing literary texts.
A more realistic evaluation is proposed by \citet{sims-etal-2019-literary}, where systems are evaluated on LitBank, a corpus containing the first 2000 tokens of 100 different books collected from Project Gutenberg.
The evaluation does not assume access to a predefined character list, matching a more realistic external validation of downstream literary analysis.
In addition, researchers may not have access to the latest GPU cards.
Thus, we report total pipeline inference times conducted on a consumer-grade GPU which does not have access to Flash-Attention or bfloat16.


LitBank contains 100 Project Gutenberg passages annotated for entities \cite{bamman-etal-2019-annotated}, coreference \cite{bamman-etal-2020-annotated} and quotation attribution  \cite{sims-bamman-2020-measuring}.
Following \citet{sims-etal-2019-literary}, we apply CoNLL F1, a standard cluster overlap metric, to the task of quotation attribution by considering a predicted cluster as the set of all quotations detected as being spoken by the same character, and gold clusters as similar sets constructed using gold attribution data.


We replace BookNLP's quotation attribution model with our best performing models (\texttt{ModernBookNLP} joint and direct, $T=2000$, $S=512$) and compare against standard BookNLP.
We use the model trained on the fourth cross-validation split of PDNC, and remove 10 LitBank passages that were drawn from the PDNC training split to ensure fair evaluation.
We also report coreference resolution CoNLL F1 score as the BookNLP pipeline uses attributed speakers as an additional signal to restrict spurious coreference clusters.
We report our results in Table~\ref{tab:booknlp_conll_time}.

Our proposed joint scoring approach obtains almost 10 points increase in CoNLL F1 attribution scores over standard BookNLP, and similar processing times on a consumer-grade GPU, despite ModernBookNLP using a model $4\times$ larger.
In contrast, ModernBookNLP with direct scoring shows only a modest performance increase (about 4 points), and processes novels twice as slowly as BookNLP.
For both models, the impact on coreference resolution is minimal, with a very small increase in CoNLL F1 score, indicating no substantial gains from better quotation attribution.
In Appendix~\ref{app:oracle_coref}, we discuss results with oracle coreference along with an analysis of predicted mention types, showing that ModernBookNLP is less prone to upstream coreference errors as it largely predicts named- rather than pronominal mentions (64\% vs 23\% and 13\% vs 60\% for joint scoring vs. BookNLP respectively).

\section{Conclusion}

We propose a new formalization of the quotation attribution task inspired by the span-based joint scoring paradigm.
Our joint scoring model achieves SOTA accuracy on PDNC, outperforming direct scoring and Llama-3-8b, while highlighting the benefits of larger contexts and coreference-based mentions, and processing novels more efficiently.
Through a detailed analysis of models' representations, we attribute the advantage of joint scoring over direct scoring to its ability to better cluster coreferent character mentions together. 
In real-world scenarios, ModernBookNLP highlights a clear advantage of joint scoring over direct scoring and vanilla BookNLP, providing a fast and accurate path towards large automatic literary analysis.
We believe a systematic evaluation of how off-the-shelf modules such as coreference resolution impact downstream quotation attribution is a promising direction for future work, along with designing fallback strategies in cases where quotations do not exhibit contextual mentions referring to their speaker.


\section{Limitations}
\label{sec:lim}

The principal limitation of this work is its evaluation on classical English novels, included in PDNC and LitBank.
Thus, it is unclear how our results hold for different languages and different dialogue conventions that contemporary fiction may imply.
Besides, although PDNC contains a large number of annotated quotations, the total number of available books is rather small, limiting the diversity of narrative styles and may make the estimated robustness of the model sensitive to the particular works included in the benchmark.

Another limitation lies in the standard evaluation setup of PDNC.
Our strongest systems rely on upstream coreference clusters produced by BookNLP and, in the PDNC experiments, on resolving these clusters against annotated character lists.
Errors or omissions in mention detection, coreference resolution, or character-list construction may propagate to quotation attribution, especially in real-world scenarios where gold character-lists are unavailable.
Indeed, while our evaluation of the best performing model on such real-world scenarios still shows substantial improvements over standard BookNLP, it is far from showing perfect performance.
We believe a systematic evaluation of how off-the-shelf quotation detection, entity detection, and coreference resolution impact downstream quotation attribution is an interesting direction for future work.

In addition, our newly proposed joint scoring formalization still relies on upstream coreference resolution, and does not provide fallback solutions in cases where ground-truth mentions might be unavailable in the considered context.
We found that this situation only happens for $2\%$ of PDNC quotations when using a context of $2,000$ tokens (which is quite low compared to the 20\% reported by \citet{michel-etal-2024-improving}, but it might be different in real-world scenarios where coreferent clusters can not be evaluated. 

Finally, our efficiency measurements are hardware dependent. In particular, PDNC inference statistics are measured on an A100 GPU using Flash Attention and bfloat16.
While we still provide an efficiency measurement for older GPUs, speedups may differ for users who only have access to CPUs.

\bibliography{anthology_aa, custom, bilbio, anthology_ab}

\appendix

\section{Restricting Coreference Clusters}
\label{app:coref}

Following standard evaluation of quotation attribution on PDNC \cite{Vishnubhotla2023, michel-etal-2024-improving}, we use the coreference clusters of \citet{Vishnubhotla2023} which use the gold character list contained within PDNC annotations to restrict potential spurious coreference clusters predicted by BookNLP.
Other evaluation protocols which do not use coreference clusters still use the gold character list as input to their model \cite{Su2023, michel-etal-2025-evaluating}.
In particular, given an output mention cluster $\mathbf{M}_c$, these mentions are kept as candidate if and only if they can be resolved to a unique annotated character in PDNC.
The matching between mention clusters and annotated characters is done by iterating through the list of aliases per character, and clusters $\mathbf{M}_c$ are removed if at least two aliases referring to different characters are present within $\mathbf{M}_c$.

While this scenario has been a standard PDNC evaluation protocol, it is slightly unrealistic as, in a real-world scenario, only predicted list of character aliases can be extracted from input novels: for example, using the Character Name Clustering module of BookNLP\footnote{\url{https://github.com/booknlp/booknlp/blob/main/booknlp/english/name_coref.py}}.
Thus, we supplement our analysis with an end-to-end evaluation of our model on LitBank.

\begin{table*}[t!]
\centering
\small
\begin{tabular}{l c c c c c c c}
\toprule
Model & (T,S) & $M_{type}$ & Overall & Non-Explicit & Explicit & Anaphoric & Implicit \\
\midrule
Direct & (1000, 256) & Alias & 85.0\tiny{$\pm$5.5} & 77.9\tiny{$\pm$6.0} & 96.4\tiny{$\pm$1.9} & 79.2\tiny{$\pm$6.1} & 76.4\tiny{$\pm$6.0} \\
Joint & (1000, 256) & Alias & 88.3\tiny{$\pm$4.3} & 83.2\tiny{$\pm$5.0} & 96.7\tiny{$\pm$1.8} & 85.3\tiny{$\pm$4.0} & 81.1\tiny{$\pm$5.8} \\
\midrule
Direct & (500, 256) & Coref & 88.9\tiny{$\pm$2.4} & 83.8\tiny{$\pm$2.6} & 96.6\tiny{$\pm$1.8} & 82.8\tiny{$\pm$1.4} & 85.1\tiny{$\pm$3.9} \\
Direct & (1000, 256) & Coref & 90.7\tiny{$\pm$2.4} & 86.6\tiny{$\pm$2.8} & 96.9\tiny{$\pm$1.9} & 86.5\tiny{$\pm$2.8} & 86.9\tiny{$\pm$5.2} \\
Direct & (2000, 512) & Coref & 90.2\tiny{$\pm$3.1} & 86.3\tiny{$\pm$3.8} & 96.1\tiny{$\pm$1.3} & 86.0\tiny{$\pm$2.3} & 85.7\tiny{$\pm$6.0} \\
\midrule
Joint & (500, 256) & Coref & 90.8\tiny{$\pm$2.2} & 86.7\tiny{$\pm$2.3} & 96.7\tiny{$\pm$1.8} & 86.6\tiny{$\pm$1.4} & 87.5\tiny{$\pm$4.0} \\
Joint & (1000, 256) & Coref & 91.7\tiny{$\pm$2.0} & 88.1\tiny{$\pm$2.6} & 97.0\tiny{$\pm$1.5} & 89.5\tiny{$\pm$2.1} & 87.8\tiny{$\pm$4.2} \\
Joint & (2000, 512) & Coref & \textbf{92.3}\tiny{$\pm$2.3} & \textbf{89.2}\tiny{$\pm$2.6} & \textbf{97.1}\tiny{$\pm$1.5} & \textbf{90.5}\tiny{$\pm$1.5} & \textbf{88.6}\tiny{$\pm$4.6} \\
\bottomrule
\end{tabular}
\caption{Quotation attribution accuracy on PDNC across context sizes ($T$), mention types and downstream performance for minor characters.}
\label{tab:add_minor}
\end{table*}

\section{Training and Inference details}
\label{app:app_1}

\paragraph{Data Creation Pipeline}

To build the data inputs for joint scoring, we build context segments of length $T$ and stride $S$.
We ensure that context boundaries are split at the sentence level.
Thus, as some quotations might sometimes be cut at the end of a contextual segment, we use striding to ensure that these split quotations are still fully integrated in a subsequent segment, and do not integrate the split quotation as part of the predictions in the initial segment.
As a result of striding, some quotations might receive two predictions if they fall in two consecutive segments. In such cases, we always keep the prediction from the first context segment as the final prediction.

To ensure fair comparison between joint and direct scoring methods, we build direct scoring inputs by extracting all quotations $q$ within a joint scoring contextual segment $\mathcal{C}$, and use $\mathcal{C}$ as the input context for all quotations $q$ within it.
Similar to above, we only use the first occurrence of a quotation (without considering split quotations) to convert it into a direct scoring input.
Note that this setup is used to ensure fair comparison across systems, but that standard approach of quotation attribution would instead use different contextual windows for each quotation as described in Section \ref{sec:problem}.

\paragraph{Baselines} We note that reported baseline results from Table~\ref{tab:results} are directly taken from their respective article, except for the Llama-3-8b based approach, from which we gathered the predictions and calculated a 5-fold cross-validation accuracy with PDNC splits.
Thus, each reported score is directly comparable to our direct and joint scoring results since reported baselines all use the same evaluation protocol (5-fold cross-validation splits with PDNC splits) and information gold aliases list of PDNC characters. 
In addition, Llama-3-8b inference times reported by \citet{michel-etal-2025-evaluating} (from 600 to 3600 seconds) were evaluated using an 8-bits quantized model on a similar GPU infrastructure (single A100 GPU), and are thus directly comparable.

\paragraph{Hyperparameters} 
For training, we set the number of layers of the MLP $\psi$ to $2$ with a hidden size of $d^{BERT}\times2$ where $d^{BERT}=1024$ is the hidden size of \texttt{ModernBERT-large}. We use a learning rate of 7e-5 with the Adam optimizer, and train for 10 epochs, gathering test accuracy using the model checkpoint from the epoch that achieved best validation accuracy.
For direct scoring models, we use a batch size of 32 quotations, while we use a batch size of 16 contexts for joint scoring systems.

\paragraph{Minor Characters}
For PDNC, we follow standard practices and report scores when removing minor characters, that is characters that are mentioned less than 10 times \cite{vishnubhotla-etal-2022-project, Vishnubhotla2023}. These represent approximately 9000 quotations.
Note that we included quotations of minor characters during training of all our models, but only removed them when reporting results in Section~\ref{sec:quant}.
Thus, we report quotation attribution accuracy for minor characters in Table~\ref{tab:add_minor}.
We observe similar trends as for the main characters, where joint scoring still dominate scores. However, for minor characters, the alias-based models display smaller gaps with the coreference-based results, which is likely due to upstream coreference resolution errors on such minor characters.

\section{Q-M Separability and Quotation Attribution}
\label{app:corr_qa}
\begin{figure}
    \centering
    \includegraphics[width=\linewidth]{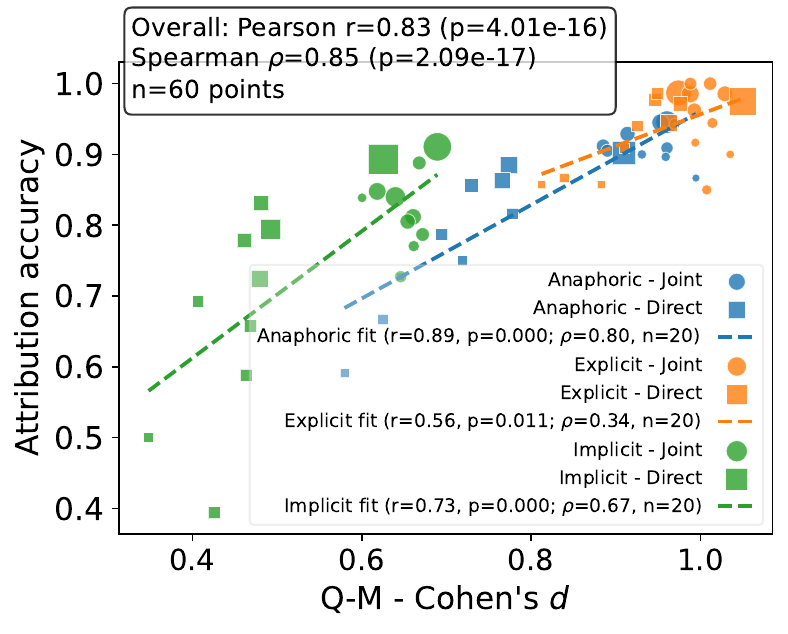}
    \caption{Attribution accuracy vs. Q-M separability (Cohen's d) per token-distance bin. Each point is one (model, quote type, distance bin) combination; marker shape encodes model (\textit{circles}=Joint, \textit{squares}=Direct), colour encodes quote type, and larger markers indicate more distant bins. Dashed lines show a per-quote-type linear fit. Accuracy and separability are strongly correlated overall, especially within Anaphoric and Implicit quotations individually.}
    \label{fig:corr_sep}
\end{figure}

To test whether our stated hypothesis are also predictive of downstream attribution accuracy, we proceed to the following analysis: for each test quotation we gather its predicted mention, and find the distance bin in which it falls into (we consider similar distance bins up to 2000 tokens as above). Then, for each distance bin, model and quotation type, we gather average attribution accuracy and Quotation-Mention (Q-M) Cohens' $d$. The rationale behind this choice is that, when a model predicts a mention within a distance bin, the local mention distractors can affect model decisions ; and we hypothesize that stronger Q-M separability can affect downstream performance in this case.
Then, for each quotation type, we gather (distance bin, model accuracy, Cohen's $d$) points and fit linear regressions.
Figure~\ref{fig:corr_sep} displays the results.
Overall, we observe a strong, highly significant positive relationship (Spearman $\rho=0.85$), confirming that Q-M separability is a reliable proxy for accuracy.
It is notably the case for Anaphoric ($\rho=0.8$) quotations, consistent with our finding that joint scoring's accuracy advantage over direct scoring is largest for this category and is attributed to preserved Q-M separation at long range \textbf{H2.}
Similarly, Implicit quotations exhibit high correlation with attribution accuracy ($\rho=0.67$).
Here, model comparison is particularly interesting: direct scoring shows systematically lower separability and accuracy, while joint scoring show consistent separability and large accuracies.
Explicit quotations, by contrast, show a weaker and non-significant within-type relationship.
We interpret this as a ceiling effect: both separability and accuracy are already saturated for Explicit quotations across nearly all conditions.

Taken together, these results reinforce the representation-level account of joint scoring's advantage: attribution accuracy is found highly impacted by how well a model's representations separate a quotation's true speaker from distractor mentions.
Separability is precisely what joint scoring preserves more effectively than direct scoring at long range for the harder, non-explicit quotation types, explaining joint scoring's advantage especially on anaphoric quotations which were found more correlated with separability.

\section{LitBank Evaluation with Oracle Coreference}
\label{app:oracle_coref}

\begin{table*}[t!]
\centering
\small
\setlength{\tabcolsep}{4.5pt}
\begin{tabular}{lcccc|cccc}
\toprule
& & & & & \multicolumn{4}{c}{Predicted $M_{type}$} 
\\ 
\cmidrule(lr){6-9}
& \textbf{MUC} & \textbf{B$^3$} & \textbf{CEAF$_{\phi_4}$} & \textbf{CoNLL F1} & PROPN & PRON & DET & NOUN\\
\midrule
BookNLP                  & 87.67 & 75.45 & 70.09 & 77.74 & 23\% & 60\% & 12\% & 3\% \\
\texttt{ModernBookNLP} (Direct)    & 88.01 & 75.15 & 69.31 & 77.49 & 54\% & 22\% & 15\% & 6\%  \\
\texttt{ModernBookNLP} (Joint)     & \textbf{90.10} & \textbf{80.44} & \textbf{73.22} & \textbf{81.25} & 64\% & 13\% & 13\% & 7\%  \\
\bottomrule
\end{tabular}
\caption{F1 score for cluster overlap metrics on LitBank's quotation attribution when using oracle coreference resolution instead of predicted coreference clusters, and proportion of predicted mention types ($M_{type}$). ModernBookNLP with joint scoring still improves over BookNLP, but with lower magnitude than with predicted coreference clusters, which is largely due to its larger proportion of predicted named mentions.}
\label{tab:f1_scores}
\end{table*}

Following \citet{sims-bamman-2020-measuring}, we analyse an \textit{oracle version} of BookNLP, which replaces the upstream coreference resolution step by the manually annotated coreference clusters of LitBank \cite{bamman-etal-2020-annotated}.
We report results in Table~\ref{tab:f1_scores}. Interestingly, BookNLP is the one benefiting the most of using oracle coreference clusters (around 9 CoNLL F1 points), reaching direct ModernBookNLP performance.
Indeed, both variants of ModernBookNLP show smaller improvements of 4 to 5 CoNNL F1 points.

To understand the reason why, we calculate the proportion of predicted mention types, using BookNLP predicted Part of Speech tags, which are also reported in Table~\ref{tab:f1_scores}.
We see that both direct and joint ModernBookNLP attribute quotations to named-mentions (PROPN) 54\% and 64\% of the time respectively, against only 23\% for standard BookNLP.
In contrast, the latter mostly attributes quotations to pronouns (PRON), with 60\% of predicted mention types as pronouns, against 22\% and 13\% for direct and joint scoring.
Since pronouns are the type of mentions that are most likely to be wrongly resolved by coreference resolution, BookNLP benefit substantially from oracle coreference resolution.
In contrast, ModernBookNLP mostly associates quotes with named-mentions, which are less likely to be wrongly resolved, and thus benefiting less from coreference resolution. 

This result is interesting: from our main results on PDNC (Section \ref{sec:quant}, Table~\ref{tab:results}), we found a large performance gap between alias-based and coreference-based attribution systems at similar context length.
However, we still observe that coreference-based attribution systems largely predict unambiguous named-mentions as candidate speakers although they could theoretically predict pronominal and nominal mentions.
This suggest that even when predicting an anaphoric quotations referred to with pronouns (``she said''), our coreference-based systems will likely prefer to attribute the quotations to a named-mention than its direct, pronominal contextual evidence (the pronoun ``she'').
A potential interpretation lies in how coreference information is integrated within the attribution objectives $L(q)$ and $L_{joint}$.
Since all mention candidates referring to a quote's speaker are seen as positive examples during training (whether jointly or not), models naturally learn to resolve coreference information through the attribution task.
Besides, our results of Section~\ref{sec:rsa} show that such signal is already present in ModernBERT, and is better preserved by joint scoring than direct scoring.

Thus, in our formalizations, well-trained coreference-based attribution models might understand the underlying per-speaker coreferent clusters (\textit{i.e.} the set of all mentions, regardless of their type, referring to this speaker), and choose to predict a named-mention rather than a pronominal or nominal mention as named-mentions are less ambiguous and likely to contain stronger predictive signal for quotation attribution.
This departs from standard BookNLP which uses a quote-masking strategy during training to force its quotation attribution system to predict contextual evidence (the ``she'' in ``she said''), which in turn is more prone to off-the-shelf coreference prediction errors.


\end{document}